%% file: 0-main.tex
\PassOptionsToPackage{table}{xcolor}
\documentclass[sigconf]{acmart}

\AtBeginDocument{%
  }

\copyrightyear{2026}
\acmYear{2026}
\setcopyright{cc}
\setcctype{by}
\acmConference[HCOMP 2026]{2026 ACM Conference on Human-AI Complementarity and Alignment}{September 27--30, 2026}{Alexandria, VA, USA}
\acmBooktitle{2026 ACM Conference on Human-AI Complementarity and Alignment (HCOMP 2026), September 27--30, 2026, Alexandria, VA, USA}
\acmDOI{10.1145/3834580.3838749}
\acmISBN{979-8-4007-2894-5/2026/09}

\usepackage{subcaption}
\usepackage{multirow}
\usepackage{colortbl}
\usepackage{makecell}
\usepackage{listings}

\usepackage{rotating}
\usepackage{xspace}
\usepackage{cleveref}
\usepackage{soul} 

\input{commands}

\begin{document}

\title{Information Satisfaction: A Reader-Centered Axis for Summarization Evaluation}

\author{Isabel Cachola}\affiliation{\institution{St. Edward's University}\city{Austin}\state{TX}\country{USA}}\authornote{Work conducted at Johns Hopkins University. Email: icachola@stedwards.edu}
\author{William Walden}\affiliation{\institution{Johns Hopkins University}\institution{Human Language Technology Center of Excellence}\city{Baltimore}\state{MD}\country{USA}}
\author{Reno Kriz}\affiliation{\institution{Johns Hopkins University}\institution{Human Language Technology Center of Excellence}\city{Baltimore}\state{MD}\country{USA}}
\author{Mark Dredze}\affiliation{\institution{Johns Hopkins University}\city{Baltimore}\state{MD}\country{USA}}


\begin{abstract}
The majority of work on summarization evaluation focuses on general summary quality (e.g., ROUGE, BERTScore) or specific desired properties (e.g., readability, factuality). However, these metrics fail to measure the utility of a summary to an individual user. For example, a biomedical researcher learning about the latest vaccine research will have different informational needs from a family doctor. Query-focused summarization captures part of this need, but in practice, users rarely state everything relevant in a query: a single short query is likely inadequate to distinguish the needs of a researcher from those of a physician. By contrast, a reader's background or \emph{persona}---their role and expertise---is comparatively stable across queries and recovers much of this missing context, which makes it a practical signal for assessing whether a summary satisfies that reader's needs. In this work, we assess how sensitive popular summarization metrics are to both informational and persona differences, and find that many popular metrics, including strong LLM-as-judge metrics, fail basic perturbation tests of informational content. We additionally conduct an expert human evaluation, measuring summary preferences based on information satisfaction given a specific person's background and use case. We find that both traditional and LLM-based metrics are insufficient measures of information satisfaction and agree poorly with human judgment.
\end{abstract}

\ccsdesc[500]{Information systems~Summarization}
\ccsdesc[300]{Human-centered computing~User studies}
\ccsdesc[100]{Computing methodologies~Natural language processing}
\ccsdesc[100]{Computing methodologies~Machine learning}

\keywords{summarization evaluation, query-focused summarization, information satisfaction, reader personas, evaluation metrics, LLM-as-judge, human evaluation, metric correlation, user-centered NLP}


\maketitle

\input{1-Introduction}
\input{2-Related-Works}
\input{3-Experimental-Setup}
\input{4-Human-Evaluation-Setup}
\input{5-Robustness-Testing-Results}

\input{6-Human-Evaluation-Results}

\input{7-Conclusion}


\bibliographystyle{ACM-Reference-Format}
\bibliography{custom}

\appendix

\input{8-appendix}

\end{document}

%% file: commands.tex
\newcommand{\badge}[2][red]{%
  {\setlength{\fboxsep}{2pt}%
   \colorbox{#1!60!black}{\hspace{1pt}\textcolor{white}{#2}\hspace{1pt}}}}

\newcommand{\chsevenrqone}{\small \badge[blue]{RQ1}\xspace}
\newcommand{\chsevenrqtwo}{\small \badge[orange]{RQ2}\xspace}

\makeatletter
\newcommand\dash@unit{\hbox to 3.5pt{\hss\vrule\@width.5pt\@height\arrayrulewidth\hss}}
\def\hdash#1#2{%
  \omit
  \@multicnt#1%
  \advance\@multispan\m@ne
  \ifnum\@multicnt=\@ne\@firstofone{&\omit}\fi
  \@multicnt#2%
  \advance\@multicnt-#1%
  \advance\@multispan\@ne
  \leaders\dash@unit\hfill
  \cr
  \noalign{\vskip-\arrayrulewidth}}
  \definecolor{refbased}{rgb}{0.85, 0.92, 0.98}    
\definecolor{refadj}{rgb}{0.88, 0.97, 0.88}       
\definecolor{reffree}{rgb}{0.98, 0.95, 0.80}       

\definecolor{promptbg}{gray}{0.93}

\lstdefinestyle{promptstyle}{
    backgroundcolor=\color{promptbg},
    basicstyle=\scriptsize\ttfamily,
    breaklines=true,
    breakatwhitespace=false,
    frame=none,
    aboveskip=0pt,
    belowskip=0pt,
    xleftmargin=6pt,
    xrightmargin=6pt,
    keepspaces=true,
    columns=flexible,
}
\makeatother

\newcolumntype{C}[1]{>{\centering\arraybackslash}p{#1}}

\definecolor{matchTrue}{RGB}{217,237,217}    
\definecolor{matchFalse}{RGB}{248,215,218}   

%% file: 1-Introduction.tex
\section{Introduction}

The value of a summary is rarely intrinsic; rather, it is realized only when the summary helps a particular reader accomplish a particular goal. Yet automatic summarization has long been evaluated through metrics that aim to capture a general notion of quality as a property of the summary itself. Lexical overlap measures like ROUGE compare candidate summaries against reference texts~\cite{lin2004rouge}; embedding-based approaches like BERTScore assess semantic similarity~\cite{zhang2020bertscore}; and more targeted metrics evaluate specific dimensions such as factual consistency, coherence, or readability~\cite{kryscinski2019neural, wang2020asking, fabbri2021summeval, luo2022readability}. More recently, Large Language Models (LLMs) have been deployed as judges, promising more nuanced assessments that approach human-level evaluation~\cite{zheng2023judging,liu2023geval}. Across this proliferation of metrics, however, quality is treated as a fixed attribute that can be measured against a gold standard or set of reference criteria, independent of who is reading.

This framing obscures a fundamental truth about summarization: a summary is only useful insofar as it serves the informational needs of the person reading it. Consider a recent paper on mRNA vaccine development. A biomedical researcher may need a summary that preserves methodological detail, statistical results, and connections to prior literature, while a family physician reading the same paper may be best served by a summary emphasizing clinical implications, patient guidance, and contraindications. The source document is the same in each case, and a reference summary written for one audience may score highly under standard metrics while wholly failing to satisfy another reader's needs. Most existing evaluation paradigms have little to say about this mismatch because they either do not model the reader at all~\cite{lin2004rouge, zhang2020bertscore} or focus on stylistic preferences of a reader, rather than informational~\cite{liu2023geval,wu-etal-2025-seeval}. 

To capture this reader-centered notion of utility, we introduce \textit{information satisfaction} as an axis of summarization evaluation. Information satisfaction measures the extent to which a summary resolves the specific informational need that motivated the reader to consult the source in the first place---operationalized in this work as a query paired with a \emph{persona} describing the reader's role and expertise. In theory, a sufficiently detailed query could encode everything about the reader, negating the need for the persona. In practice, however, readers rarely provide such information in their requests, leaving the reader's background, expertise, and purpose implicit. A persona recovers exactly this missing context, and because a reader's role and expertise are comparatively stable across the many queries they issue, it can be specified once and reused rather than re-elicited for every request. We therefore treat the query and the persona as complementary, with the query expressing the immediate question and the persona supplying the relatively static reader context that the query leaves unsaid.
A summary can be fluent, factually accurate, and a faithful condensation of its source while still leaving the reader's needs unaddressed. Conversely, a stylistically rough summary may fully satisfy the query at hand for a reader with the right background. By foregrounding both the query and the reader behind it, information satisfaction reframes evaluation around the reader's purpose rather than the summary's surface properties.

Given this reframing, we ask whether existing summarization metrics are equipped to measure information satisfaction. We organize our investigation around two research questions:

\paragraph{{\chsevenrqone\label{chsevenrq1}} Are evaluation metrics sensitive to variation in informational content and reader persona?}
A metric that meaningfully tracks information satisfaction should respond when the information in a summary shifts relative to what the reader needs—either because the summary's content changes or because the persona consuming it changes. We probe this property by systematically varying both axes and measuring how popular automatic metrics, including traditional reference-based measures and LLM-as-judge approaches, respond to the perturbations.

\paragraph{{\chsevenrqtwo\label{chsevenrq2}} Do evaluation metrics agree with human judgments of information satisfaction?} 
Even a metric that is sensitive to content and persona changes is only useful if its judgments align with those of real readers. We conduct an expert human evaluation in which annotators rate summaries with respect to a specified query and persona, and we measure the correlation between these human judgments and the scores produced by automatic metrics.

Here, we show that many popular metrics, including strong LLM-as-judge metrics, fail simple perturbation tests to the informational content of a summary. It is thus unsurprising that we further observe that the same metrics show poor agreement with human judges of information satisfaction. Taken together, our results suggest that neither traditional nor LLM-based metrics adequately capture information satisfaction, and that progress on user-centered summarization requires evaluation frameworks that explicitly model who the summary is for and what they seek to learn.

%% file: 2-Related-Works.tex
\section{Related Works}\label{sec:ch-7-related_work}

\textit{Automatic summarization evaluation} has historically centered on $n$-gram overlap with reference summaries, most notably ROUGE~\cite{lin2004rouge}. While simple and reproducible, ROUGE rewards surface overlap rather than semantic equivalence and correlates only weakly with human judgments of quality~\cite{liu2008correlation,bhandari2020reevaluating}. Embedding-based metrics such as BERTScore~\cite{zhang2020bertscore} and MoverScore~\cite{zhao2019moverscore} address this by comparing contextualized representations, and learned metrics like BLEURT~\cite{sellam2020bleurt} train directly on human ratings. A parallel line of work targets specific aspects of summary quality, particularly factual consistency, with metrics including FactCC~\cite{kryscinski2020evaluating}, QAGS~\cite{wang2020asking}, QuestEval~\cite{scialom2021questeval}, and SummaC~\cite{laban2022summac}. More recently, large language models have been used directly as evaluators of summary quality and have been shown to outperform earlier automatic metrics on several aspects of summary evaluation~\cite{liu2023geval,fu2024gptscore,chiang2023llmeval}. These metrics share a common assumption that summary quality can be assessed independently of the reader, which is precisely the assumption we challenge.

\textit{Query-focused summarization (QFS)} generates summaries tailored to a user-provided query rather than producing a generic synopsis~\cite{daume2006bayesian,dang2005duc}, with approaches ranging from early extractive methods that scored sentences by query relevance~\cite{wan2007manifold,otterbacher2009biased} to neural systems that generate query-conditioned abstractive summaries~\cite{baumel2018query,xu2020coarse,vig2022exploring,gantt-etal-2024-event}, supported by datasets such as QMSum~\cite{zhong2021qmsum} and AQuaMuSe~\cite{kulkarni2020aquamuse}. A parallel line of work conditions summarization on reader attributes, including controllable systems that specify length, style, or focus~\cite{fan2018controllable,he2022ctrlsum}, audience-adaptive approaches targeting populations such as children, experts, or laypeople~\cite{chandrasekaran2020overview,august2022paperplain,goldsack2022making,luo2022readability}, and persona-based prompting of large language models~\cite{deshpande2023toxicity,zhang2024personallm}. However, QFS evaluation has largely relied on reference-based metrics applied against query-specific summaries~\cite{zhong2021qmsum,vig2022exploring}; our work extends this framing by pairing each query with an explicit reader persona and interrogating whether existing metrics can detect query-conditioned differences in informational content.

\textit{Human Evaluation} A number of datasets have been released with human annotations of summary quality, enabling meta-evaluation of automatic metrics. SummEval~\cite{fabbri2021summeval} provides expert ratings of system outputs on CNN/DailyMail across coherence, consistency, fluency, and relevance. FRANK~\cite{pagnoni2021frank} contributes fine-grained factual error annotations, and~\citet{tang2022understanding} extends this with error annotations across additional summarizers and datasets. More recent benchmarks elicit preference judgments via pairwise comparison of system outputs~\cite{goyal2022news,liu2023benchmarking}. Across these resources, annotations are collected without reference to a particular reader's goals; our evaluation anchors judgments to a specified persona and query, enabling direct measurement of information satisfaction.

%% file: 3-Experimental-Setup.tex
\section{Experimental Setup \hyperref[chsevenrq1]{\chsevenrqone}}\label{sec:ch-7-experimental-setup}

We evaluate the robustness of automatic summarization metrics by
applying controlled perturbations to reference summaries and measuring
whether each metric responds in the expected direction. 

We draw samples from 4 scientific summarization corpora that span
a range of domains, document lengths, and intended audiences.
\textbf{arXiv} and \textbf{PubMed} \citep{cohan2018discourse} provide
long-document scientific articles paired with their abstracts, while
\textbf{SciTLDR} \citep{cachola2020tldr} contributes extreme,
single-sentence TL;DR summaries of Computer Science papers. \textbf{eLife} contains editor-written lay summaries of biomedical articles~\cite{goldsack2022making}.
For each dataset, we use the test set and subsample $N{=}50$ documents for experimentation.

We evaluate a broad set of metrics covering the major families used in
the summarization literature. For lexical overlap with a reference
summary, we include ROUGE-1/2/L \citep{lin2004rouge},
BLEU \citep{papineni2002bleu}, chrF \citep{popovic2017chrf}, and
METEOR \citep{banerjee2005meteor}. We additionally report a set of
surface-level statistics of the summary itself, computed via
DataStats \citep{grusky2018newsroom}: extractive coverage, extractive
density, compression ratio, the percentage of novel 1/2/3-grams
relative to the source, and summary length. A spaCy-based
\citep{honnibal2020spacy} syntactic-complexity metric, measured in
both words and sentences, is included as a reference-free baseline
that captures only surface properties of the summary. We use the
SumEval package to compute the above metrics~\cite{fabbri2021summeval}.
 
For non-lexical, embedding- and model-based evaluation, we include
BERTScore \citep{zhang2020bertscore}, which computes token-level
cosine similarity using contextual embeddings.\footnote{We use the default
RoBERTa-large model for BERTScore.} We further include the reference-free metrics
SUPERT \citep{gao2020supert}, SummaQA \citep{scialom2019summa}, and
BLANC \citep{vasilyev2020blanc}, which score summaries against the
source document only.
 
For LLM-based evaluation, we run each metric under two judges, a
larger Llama-3.3-70B model\footnote{\texttt{meta-llama/Llama-3.3-70B-Instruct}} and a smaller Prometheus-7B
model\footnote{\texttt{prometheus-eval/prometheus-7b-v2.0}}~\citep{kim2023prometheus, grattafiori2024llama3}, in order to assess judge sensitivity.
We include FActScore \citep{min2023factscore}, which verifies atomic
facts against the source. Following~\citet{liu2023geval}, we use an
LLM-as-judge metric \citep{zheng2023judging} that scores summaries
along the dimensions of relevance, consistency, fluency, coherence, and informativeness, and that also produces an overall score. Finally, we include
LLM-as-judge prompts that mirror the instructions provided to the human annotators (\emph{LLM Judge: Annotator} in our results below).

Finally, we experiment with a variant of FActScore that measures the utility of information \emph{nuggets} \citep{voorhees-2003-evaluating-answers, voorhees2003overview} to the user, rather than the fidelity of the nuggets to the source. We refer to these metrics as \emph{Persona Recall} and \emph{Persona Precision}. Similar to past work in claim decomposition~\cite{wanner-etal-2025-dndscore,jiang-etal-2025-core}, both decompose a summary into subclaims, via an LLM extraction step, then judge each nugget against the target persona (role, domain, info needs, query). Persona Precision measures the fraction of summary nuggets that an LLM judge deems relevant to the persona, penalizing off-topic or persona-irrelevant content. Persona Recall first prompts the LLM to generate a list of information requirements implied by the persona and source document, then checks what fraction of those requirements is covered by the summary's nuggets, penalizing omissions of persona-critical information. Together these metrics capture whether a summary contains the right information for a specific reader and nothing extraneous.

\subsection{Perturbation tests}
\label{sec:perturbations}

We apply five perturbations to each summary, each producing an ordered
sequence of progressively perturbed variants. For each test, we state
the \emph{expected direction} in which a faithful metric should move as the perturbation level increases.

 \textbf{Distractor sentences} (expected: decrease). Up to five
    sentences sampled uniformly at random from \emph{other} documents in
    the corpus are appended one at a time to the summary, following the
    style of distractor-injection robustness probes~\citep{kryscinski2019neural,gabriel2021goyfaithful}.

 \textbf{Incremental addition} (expected: increase). Starting
    from the empty string, we add one sentence of the original summary
    at a time until the full summary is reconstructed. This probes
    whether a metric distinguishes partial from complete summaries.

\textbf{Lengthen} (expected: stable). An LLM is prompted to
    expand the summary into longer prose without introducing new
    information; the source document is provided as grounding context.
    
\textbf{Shorten} (expected: stable). The inverse of
    \emph{lengthen}: the LLM is asked to condense the summary while
    preserving content. Together these two tests measure
    length-sensitivity, a known failure mode of overlap-based metrics
    \citep{sun2019compare}.

\textbf{Different audience} (expected: decrease). The summary
    is rewritten by the LLM for a different target audience drawn from
    \{undergraduate student, journalist, domain expert\}. This isolates
    the effect of audience-aware emphasis, which persona-aware metrics
    should detect as a quality drop relative to the original audience.

We use Llama-3.3-70B to generate the LLM-based pertubations (tests~3-5). 

We compute the Spearman rank correlation $\rho$~\cite{spearman1904proof} between perturbation level and mean metric score. We additionally report \emph{Monotonicity} (M), the fraction of consecutive transitions between perturbation levels whose sign matches the expected direction. For \emph{stable} tests, a transition is considered correct when the absolute score change is below 5\% of the baseline score.
The expected output of each statistic, along with each expected direction, is summarized in \autoref{tab:expected_directions}.

\input{tables/expected-stats}

%% file: tables/expected-stats.tex
\begin{table}[htbp]
\centering
\footnotesize
\setlength{\tabcolsep}{6pt}
\renewcommand{\arraystretch}{1.25}
\begin{tabular}{@{}l p{.22\columnwidth} p{.22\columnwidth} p{.22\columnwidth}@{}}
\diaghead(-3,1){\hskip.18\columnwidth}{\textbf{Stat.}}{\textbf{Dir.}} & \textbf{Decrease} & \textbf{Increase} & \textbf{Stable} \\ \hline
$\rho$ & $\approx -1$ & $\approx +1$ & $\approx 0$ \\
 & (strong negative) & (strong positive) &  \\ \hdash{1}{4}
\multirow{2}{*}{M} & $\approx 1.0$ & $\approx 1.0$ & $\approx 1.0$ \\
 & (every step lowers the score) & (every step raises the score) & (every step stays within $\pm 5\%$ of baseline)
\end{tabular}
\caption{We report the expected statistic values (Stat.) by the direction of change (Dir.) for Spearman rank correlation ($\rho$) and Monotonicity (M). See \S\ref{sec:perturbations} for details.}
\label{tab:expected_directions}
\end{table}

%% file: 4-Human-Evaluation-Setup.tex
\section{Human Evaluation Collection \hyperref[chsevenrq2]{\chsevenrqtwo}}\label{sec:ch-7-human-eval-setup}

We conduct a human evaluation to assess whether conditioning summary
generation on a reader's self-reported background yields summaries that
are preferred by that reader. Crucially, we measure preference with
respect to the specific annotator who provided the query and persona,
rather than aggregating judgments across annotators. We release the annotation platform and data to support future research.\footnote{\url{https://github.com/JHU-CLSP/persona-eval-metric}}

\begin{figure*}[htbp]
    \centering
    \begin{subfigure}[t]{0.3\textwidth}
        \centering
        \includegraphics[width=.95\linewidth]{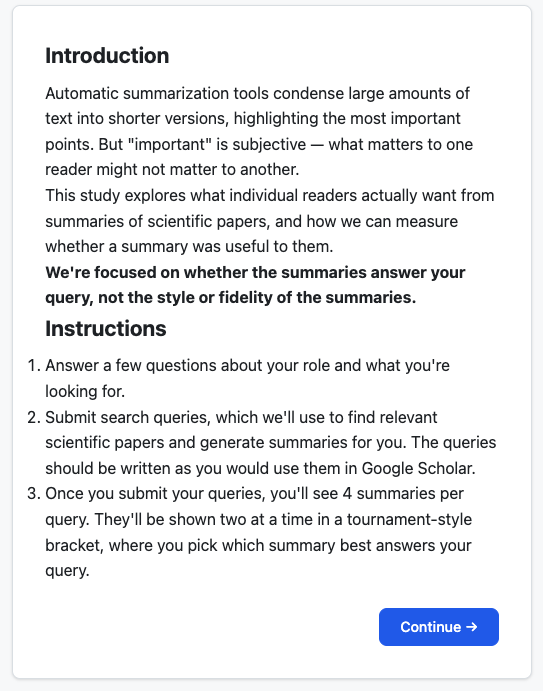}
        \caption{Introduction to the task and instructions.}
        \label{fig:interface-intro}
    \end{subfigure}
    \hfill
    \begin{subfigure}[t]{0.3\textwidth}
        \centering
        \includegraphics[width=.95\linewidth]{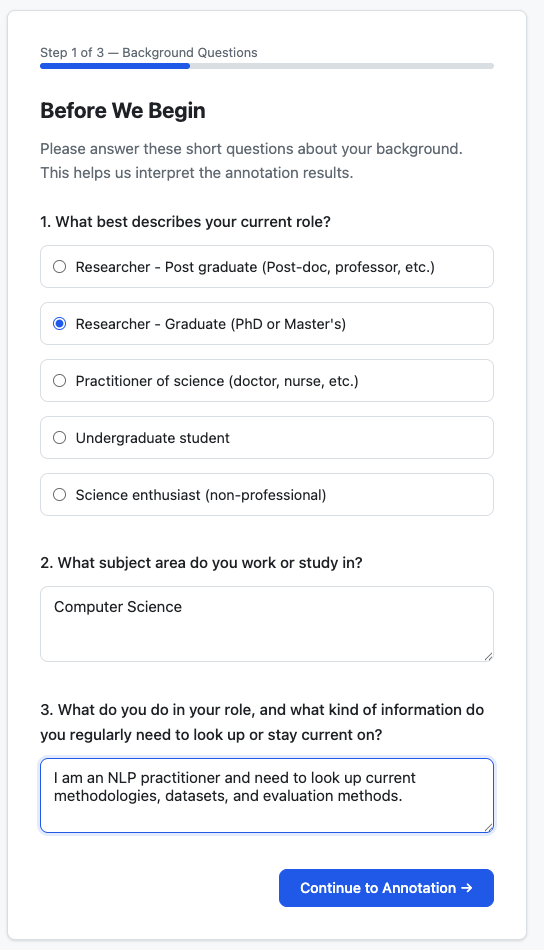}
        \caption{Profile questionnaire for building a persona.}
        \label{fig:interface-profile}
    \end{subfigure}
    \hfill
    \begin{subfigure}[t]{0.3\textwidth}
        \centering
        \includegraphics[width=.95\linewidth]{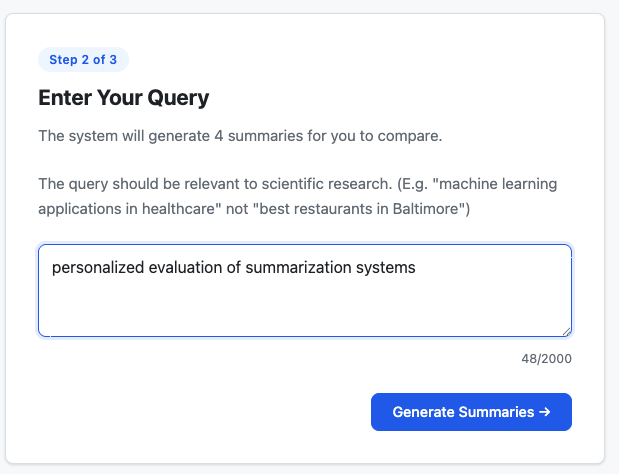}
        \caption{Annotators choose query topic.}
        \label{fig:interface-query}
    \end{subfigure}
        \vspace{1em}

    \begin{subfigure}[t]{\textwidth}
        \centering
        \includegraphics[width=.7\linewidth]{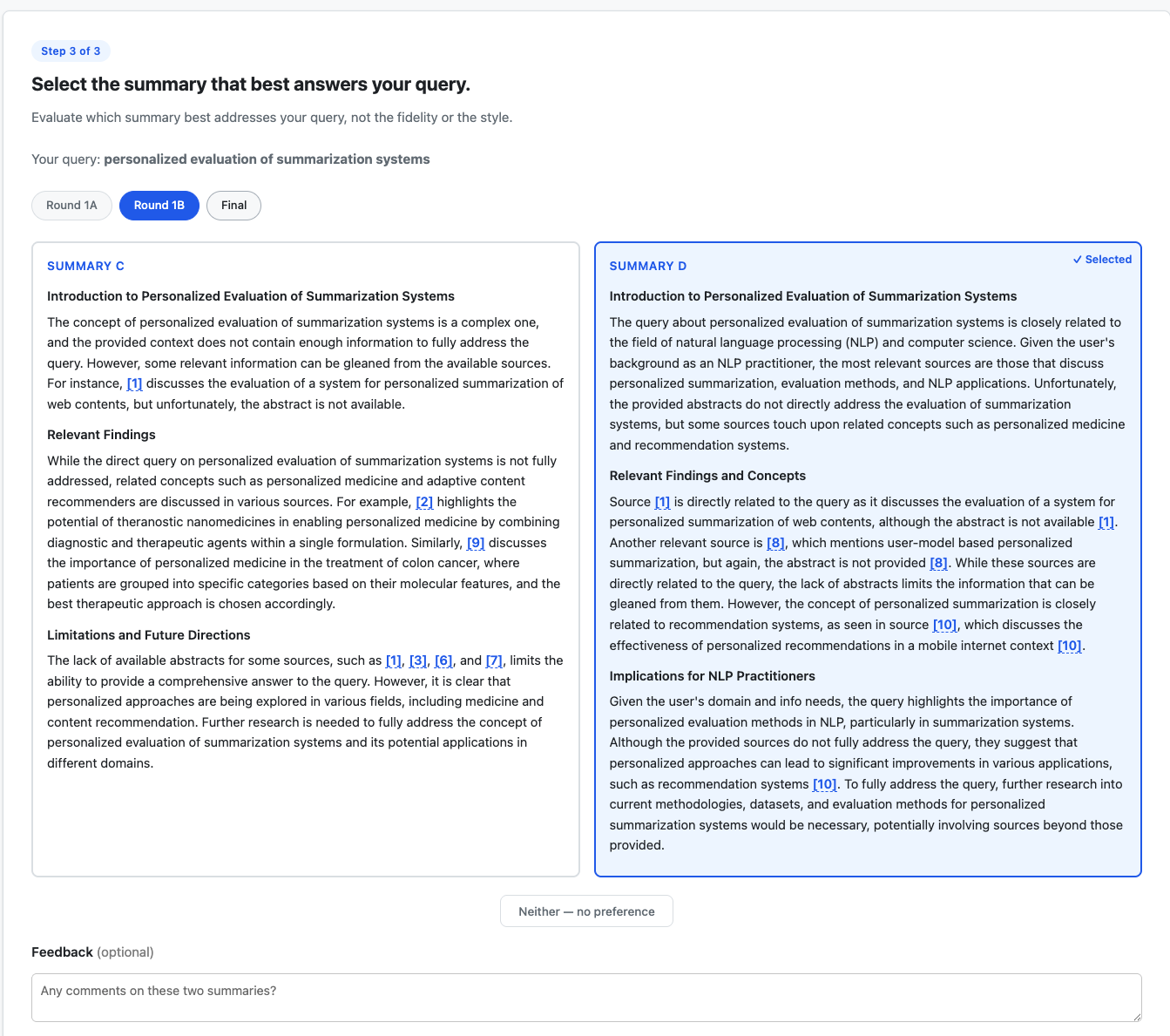}
        \caption{Example summary comparison.}
        \label{fig:interface-comparison}
    \end{subfigure}

    \caption{Annotation workflow example. Steps~\ref{fig:interface-query}-\ref{fig:interface-comparison} continue for each query assigned to the annotator.}
    \label{fig:annotation-interface}
\end{figure*}

\textit{Annotation protocol.}
Each annotator interacts with a custom web-based platform
(see~\Cref{fig:annotation-interface} for screenshots). A session proceeds in four stages:
(i)~the annotator completes a short pre-task questionnaire eliciting
self-reported expertise, research area, and use case, which together
constitute their \emph{persona profile};
(ii)~they submit a free-form natural-language query reflecting a
genuine information need from their own work;
(iii)~the system retrieves the top 10 paper abstracts matching the
query from OpenAlex\footnote{\url{https://openalex.org/}} and conditions generation on this
context; and
(iv)~the annotator ranks the resulting four summaries through a
pairwise tournament (described below). Each annotator completes this
loop for up to eight distinct queries.

\textit{Summary generation.} For each query, four candidate summaries are produced from a
$2 \times 2$ design crossing two open-weight instruction-tuned models---Llama-3.3-70B~\citep{grattafiori2024llama3}
and DeepSeek-V3.1\footnote{\texttt{deepseek-ai/DeepSeek-V3.1}}~\citep{deepseekai2024deepseekv3technicalreport}---with
two prompting conditions: one with the persona profile and one without. All
four summaries are generated from the same retrieved context (to eliminate differences arising from search performance or stochasticity) using a
shared prompt template that instructs the model to ground its answer
in the provided abstracts and to use bracketed in-text
citations~\citep{lewis2020rag,gao2023ragsurvey}.\footnote{Full prompts are
provided in appendix.}

\textit{Ranking protocol.}
Annotators rank the four summaries through a pairwise
single-elimination tournament: two pairs are compared in the first
round, and the winners from each pair advance to a final. Assignment of summaries to
bracket positions is randomized so that annotators cannot infer which
model or prompt condition produced a given summary. Pairwise
comparison has been shown to yield more reliable human judgments than
absolute Likert ratings~\citep{kiritchenko2017bws,liusie2024llmcomparative},
and the bracket structure reduces annotator effort from
$\binom{4}{2}=6$ pairwise judgments to three while producing a general
ordering over the top two summaries. For each round, the annotators also have the option to select ``neither'' if neither summary meets their information needs. If ``neither'' is selected for one or both of the first round pairs, the final round is skipped. 

\textit{Annotator recruitment.}
Standard inter-annotator agreement is not a meaningful quality signal
in our setting; information satisfaction is defined relative to a
\emph{specific} reader's background and query, and two annotators with
different backgrounds should be expected to disagree on the same
summary. Agreement among annotators would in fact suggest that our
protocol \emph{fails} to capture the persona-conditioned signal we aim to
measure. We instead ensure annotation quality through careful
recruitment. Rather than crowdsourcing, which would require
annotators to fabricate information needs, we recruit 20 volunteer
annotators directly through professional networks, all of whom hold genuine interests in supporting language technology research. Our annotator pool spans domains including computer science, medicine, and materials
science, and roles including researchers, practicing professionals
(e.g., nurses), and undergraduate students. Annotators were \emph{not} required to use all assigned queries if they did not have a genuine information need for all eight. Because summaries are
generated live in response to each annotator's own query, every
judgment reflects an authentic information need.

%% file: 5-Robustness-Testing-Results.tex
\input{tables/perturbations-rho}
\section{Perturbation Testing Results \hyperref[chsevenrq1]{\chsevenrqone}}

Results for the five perturbation tests for all metrics described in~\Cref{sec:ch-7-experimental-setup} are presented in~\Cref{tab:all_tests_combined}. For the directional tests, we consider the test ``passed'' if $|{\rho}| \geq 0.4$ in the expected direction; in other words, if the test produces at least a moderate correlation. For the stability tests, we consider the test ``passed'' if $|{\rho}| \leq 0.2$, or if the test produces a weak correlation.  For monotonicity, we consider the test ``passed'' if $M \geq 0.5$, or if the metric moves in the expected direction at least half the time. We release the code for the metric calculation, perturbation testing, and analysis to support future research.\footnote{\url{https://github.com/JHU-CLSP/persona-eval-annotation}}

\textit{Distractor sentences.} Appending unrelated sentences should cause reference-based metrics to decrease. The lexical overlap metrics (ROUGE-1/L, BLEU, chrF, METEOR, compression) behave as expected, all reaching $\rho = -1.0$ with $M = 1.0$. More notably, BERTScore, BLANC, and SummaQA fail to produce a defined correlation, and the LLM-based metrics are inconsistent across judges: under Llama-3.3-70B only FActScore ($\rho = -0.60$) and coherence ($\rho = -0.58$) pass, while under Prometheus-7B a different and largely non-overlapping subset passes (relevance, fluency, the annotator-style judge, persona recall). The metrics most often advocated as ``content-aware'' alternatives to ROUGE are thus the \emph{least} reliable detectors of obviously off-topic content.

\textit{Incremental addition.} Reconstructing the summary one sentence at a time should cause scores to increase as the summary becomes more complete. Strikingly, ROUGE-1/2/L, BLEU, chrF, METEOR, and compression all show \emph{negative} correlations between $-0.56$ and $-0.65$. Partial summaries are scored \emph{higher} than complete ones, likely an artifact of length normalization rewarding short prefixes that overlap the reference. The Llama-3.3-70B judge is the lone bright spot: every prompt-driven dimension lies between $\rho = 0.42$ and $\rho = 0.58$ in the correct direction, although low monotonicity ($M \leq 0.13$) indicates noisy scores. The Prometheus-7B judge fails on every dimension. No reference-based metric in either the lexical or non-lexical family reliably distinguishes partial from complete summaries.

\textit{Lengthen and shorten prose.} A content-preserving rewrite that changes surface length should leave a content-aware metric approximately unchanged ($\rho \approx 0$). Almost every metric fails: all overlap-based metrics, extractive coverage and density, compression, and BERTScore reach $\rho = -1.0$ on \texttt{lengthen\_prose}, with mirrored behavior on \texttt{shorten\_prose}. Most LLM Judge dimensions also collapse to $|\rho| = 1.0$, indicating they are themselves strongly length-sensitive. The few length-invariant metrics are SummaQA, SUPERT, and a small subset of LLM-based dimensions (Llama-3.3-70B relevance, consistency, and persona precision; Prometheus-7B consistency). This extends the length-sensitivity failure mode previously documented for ROUGE~\citep{sun2019compare} to BERTScore and to most LLM-judge dimensions.

\textit{Different audience.} Rewriting a summary for a different target audience should be detected as a quality drop. The most striking result is that several LLM-based metrics move in the \emph{wrong} direction: under Prometheus-7B, fluency, informativeness, overall, and FActScore all show strong positive correlations of $0.80$--$1.00$, rating audience-shifted summaries as \emph{higher} quality than the original. The lexical overlap metrics (ROUGE-1/2, chrF, METEOR, coverage, density, compression) and BERTScore sit exactly at our pass threshold ($\rho = -0.40$), giving only a weak directional signal, and ROUGE-L, BLEU, BLANC, SummaQA, and SUPERT fail outright. Under the Llama-3.3-70B judge, only the annotator-style judge passes ($\rho = -0.77$); under Prometheus, only consistency ($\rho = -0.40$) and persona recall ($\rho = -0.60$). Compared to the much stronger responses these same metrics produce under distractor injection, this confirms that existing summarization metrics are far more sensitive to surface-level corruptions than to meaningful changes in the audience the summary serves.

\textit{Summary of robustness findings.} No single metric passes all of the tests it should, and most pass only the distractor-injection test---the perturbation furthest from realistic variation. LLM-based metrics offer a partial complement, recovering sensitivity on incremental addition (where lexical metrics catastrophically fail), but they are highly inconsistent across judge models: dimensions that pass under Llama-3.3-70B routinely fail or invert under Prometheus-7B. Most importantly, every evaluated metric is approximately insensitive to, or even \emph{anticorrelated} with, audience-shift rewrites---the dimension along which informational satisfaction varies between users. These results motivate the human evaluation in~\Cref{sec:ch-7-human-eval-results}. \\

%% file: tables/perturbations-rho.tex
\begin{table*}[htbp]
\centering
\begin{subtable}[t]{0.49\linewidth}
\centering
\footnotesize
\setlength{\tabcolsep}{2pt}
\begin{tabular}{lrrrrr}
\toprule
 & \multicolumn{3}{c}{\textit{Directional}} & \multicolumn{2}{c}{\textit{Stability}} \\
\cmidrule(lr){2-4} \cmidrule(lr){5-6}
Metric & DiffAud$\downarrow$ & Distract$\downarrow$ & Increm$\uparrow$ & Lengthen$\approx$0 & Shorten$\approx$0 \\
\midrule
\multicolumn{6}{l}{\textit{Lexical}} \\
\midrule
ROUGE-1 (F1) & \cellcolor{matchTrue}$-0.40$ & \cellcolor{matchTrue}$-1.00$ & \cellcolor{matchFalse}$-0.63$ & \cellcolor{matchFalse}$-1.00$ & \cellcolor{matchFalse}$-1.00$ \\
ROUGE-2 (F1) & \cellcolor{matchTrue}$-0.40$ & \cellcolor{matchTrue}$-0.66$ & \cellcolor{matchFalse}$-0.63$ & \cellcolor{matchFalse}$-1.00$ & \cellcolor{matchFalse}$-1.00$ \\
ROUGE-L (F1) & \cellcolor{matchFalse}$0.20$ & \cellcolor{matchTrue}$-1.00$ & \cellcolor{matchFalse}$-0.63$ & \cellcolor{matchFalse}$-1.00$ & \cellcolor{matchFalse}$-1.00$ \\
BLEU & \cellcolor{matchFalse}$-0.20$ & \cellcolor{matchTrue}$-1.00$ & \cellcolor{matchFalse}$-0.64$ & \cellcolor{matchFalse}$-1.00$ & \cellcolor{matchFalse}$-1.00$ \\
chrF & \cellcolor{matchTrue}$-0.40$ & \cellcolor{matchTrue}$-1.00$ & \cellcolor{matchFalse}$-0.60$ & \cellcolor{matchFalse}$-1.00$ & \cellcolor{matchFalse}$-1.00$ \\
Extr. Coverage & \cellcolor{matchTrue}$-0.40$ & \cellcolor{matchTrue}$-0.54$ & \cellcolor{matchFalse}$0.02$ & \cellcolor{matchFalse}$-1.00$ & \cellcolor{matchFalse}$-1.00$ \\
Extr. Density & \cellcolor{matchTrue}$-0.40$ & \cellcolor{matchTrue}$-0.54$ & \cellcolor{matchTrue}$0.54$ & \cellcolor{matchFalse}$-1.00$ & \cellcolor{matchFalse}$-1.00$ \\
Compression Ratio & \cellcolor{matchTrue}$-0.40$ & \cellcolor{matchTrue}$-1.00$ & \cellcolor{matchFalse}$-0.65$ & \cellcolor{matchFalse}$-1.00$ & \cellcolor{matchFalse}$1.00$ \\
Novel 1-grams (\%) & \cellcolor{matchFalse}$0.40$ & \cellcolor{matchFalse}$1.00$ & \cellcolor{matchTrue}$0.84$ & \cellcolor{matchFalse}$1.00$ & \cellcolor{matchFalse}$1.00$ \\
Novel 2-grams (\%) & \cellcolor{matchFalse}$0.40$ & \cellcolor{matchFalse}$0.83$ & \cellcolor{matchFalse}$-0.09$ & \cellcolor{matchFalse}$1.00$ & \cellcolor{matchFalse}$1.00$ \\
Novel 3-grams (\%) & \cellcolor{matchFalse}$0.40$ & \cellcolor{matchFalse}$0.54$ & \cellcolor{matchFalse}$-0.40$ & \cellcolor{matchFalse}$1.00$ & \cellcolor{matchFalse}$1.00$ \\
Summary Length & \cellcolor{matchFalse}$0.20$ & \cellcolor{matchFalse}$1.00$ & \cellcolor{matchTrue}$1.00$ & \cellcolor{matchFalse}$1.00$ & \cellcolor{matchFalse}$-1.00$ \\
METEOR & \cellcolor{matchTrue}$-0.40$ & \cellcolor{matchTrue}$-1.00$ & \cellcolor{matchFalse}$-0.56$ & \cellcolor{matchFalse}$-1.00$ & \cellcolor{matchFalse}$-1.00$ \\
Synt. Complexity (W) & \cellcolor{matchFalse}$0.20$ & \cellcolor{matchFalse}$1.00$ & \cellcolor{matchTrue}$1.00$ & \cellcolor{matchFalse}$1.00$ & \cellcolor{matchFalse}$-1.00$ \\
Synt. Complexity (S) & \cellcolor{matchFalse}$0.20$ & \cellcolor{matchFalse}$1.00$ & \cellcolor{matchTrue}$1.00$ & \cellcolor{matchFalse}$1.00$ & \cellcolor{matchFalse}$-1.00$ \\
\midrule
\multicolumn{6}{l}{\textit{Non-lexical}} \\
\midrule
BERTScore (F1) & \cellcolor{matchTrue}$-0.40$ & --- & --- & \cellcolor{matchFalse}$-1.00$ & \cellcolor{matchFalse}$-1.00$ \\
BLANC & \cellcolor{matchFalse}$0.80$ & --- & \cellcolor{matchFalse}$-0.31$ & \cellcolor{matchFalse}$1.00$ & \cellcolor{matchFalse}$1.00$ \\
SummaQA (F1) & --- & --- & --- & \cellcolor{matchTrue}$0.00$ & \cellcolor{matchTrue}$0.00$ \\
SUPERT & \cellcolor{matchFalse}$0.50$ & --- & --- & --- & --- \\
\midrule
\multicolumn{6}{l}{\textit{LLM judge: Llama-70B}} \\
\midrule
FActScore & \cellcolor{matchFalse}$0.40$ & \cellcolor{matchTrue}$-0.60$ & \cellcolor{matchFalse}$0.38$ & \cellcolor{matchFalse}$-1.00$ & \cellcolor{matchFalse}$-1.00$ \\
Relevance & --- & \cellcolor{matchFalse}$-0.10$ & \cellcolor{matchTrue}$0.42$ & \cellcolor{matchTrue}$0.00$ & \cellcolor{matchTrue}$0.00$ \\
Coherence & \cellcolor{matchFalse}$0.77$ & \cellcolor{matchTrue}$-0.58$ & \cellcolor{matchTrue}$0.58$ & \cellcolor{matchFalse}$1.00$ & \cellcolor{matchFalse}$1.00$ \\
Consistency & --- & \cellcolor{matchFalse}$-0.13$ & \cellcolor{matchTrue}$0.42$ & \cellcolor{matchTrue}$0.00$ & \cellcolor{matchTrue}$0.00$ \\
Fluency & \cellcolor{matchFalse}$0.77$ & \cellcolor{matchFalse}$0.65$ & \cellcolor{matchTrue}$0.58$ & \cellcolor{matchFalse}$1.00$ & \cellcolor{matchFalse}$1.00$ \\
Informativeness & \cellcolor{matchFalse}$0.32$ & \cellcolor{matchFalse}$0.85$ & \cellcolor{matchTrue}$0.42$ & \cellcolor{matchFalse}$1.00$ & \cellcolor{matchFalse}$1.00$ \\
Overall & \cellcolor{matchFalse}$0.32$ & \cellcolor{matchFalse}$-0.14$ & \cellcolor{matchTrue}$0.58$ & \cellcolor{matchFalse}$1.00$ & \cellcolor{matchFalse}$1.00$ \\
Annotator & \cellcolor{matchTrue}$-0.77$ & --- & --- & --- & \cellcolor{matchFalse}$-1.00$ \\
Persona Precision & --- & --- & --- & \cellcolor{matchTrue}$0.00$ & \cellcolor{matchTrue}$0.00$ \\
Persona Recall & \cellcolor{matchFalse}$0.80$ & \cellcolor{matchFalse}$0.20$ & \cellcolor{matchTrue}$0.45$ & \cellcolor{matchFalse}$-1.00$ & \cellcolor{matchFalse}$-1.00$ \\
\midrule
\multicolumn{6}{l}{\textit{LLM judge: Prometheus-7B}} \\
\midrule
FActScore & \cellcolor{matchFalse}$1.00$ & \cellcolor{matchTrue}$-0.40$ & \cellcolor{matchFalse}$-0.36$ & \cellcolor{matchFalse}$-1.00$ & \cellcolor{matchFalse}$-1.00$ \\
Relevance & \cellcolor{matchFalse}$-0.21$ & \cellcolor{matchTrue}$-0.56$ & \cellcolor{matchFalse}$-0.20$ & \cellcolor{matchFalse}$-1.00$ & \cellcolor{matchFalse}$-1.00$ \\
Coherence & \cellcolor{matchFalse}$0.40$ & \cellcolor{matchFalse}$-0.12$ & \cellcolor{matchFalse}$-0.01$ & \cellcolor{matchFalse}$1.00$ & \cellcolor{matchFalse}$1.00$ \\
Consistency & \cellcolor{matchTrue}$-0.40$ & \cellcolor{matchFalse}$-0.32$ & \cellcolor{matchFalse}$-0.17$ & \cellcolor{matchFalse}$-1.00$ & \cellcolor{matchTrue}$0.00$ \\
Fluency & \cellcolor{matchFalse}$0.80$ & \cellcolor{matchTrue}$-0.72$ & \cellcolor{matchFalse}$0.20$ & \cellcolor{matchFalse}$1.00$ & \cellcolor{matchFalse}$1.00$ \\
Informativeness & \cellcolor{matchFalse}$0.80$ & \cellcolor{matchFalse}$0.75$ & \cellcolor{matchFalse}$-0.11$ & \cellcolor{matchFalse}$1.00$ & \cellcolor{matchFalse}$1.00$ \\
Overall & \cellcolor{matchFalse}$0.80$ & \cellcolor{matchFalse}$-0.37$ & \cellcolor{matchFalse}$-0.04$ & \cellcolor{matchFalse}$1.00$ & \cellcolor{matchFalse}$1.00$ \\
Annotator & \cellcolor{matchFalse}$0.40$ & \cellcolor{matchTrue}$-0.65$ & \cellcolor{matchFalse}$-0.00$ & \cellcolor{matchFalse}$1.00$ & \cellcolor{matchFalse}$1.00$ \\
Persona Precision & \cellcolor{matchFalse}$0.00$ & \cellcolor{matchFalse}$0.03$ & \cellcolor{matchFalse}$-0.01$ & \cellcolor{matchFalse}$-1.00$ & \cellcolor{matchFalse}$-1.00$ \\
Persona Recall & \cellcolor{matchTrue}$-0.60$ & \cellcolor{matchTrue}$-0.60$ & \cellcolor{matchFalse}$0.09$ & \cellcolor{matchFalse}$1.00$ & \cellcolor{matchFalse}$-1.00$ \\
\bottomrule
\end{tabular}
\caption{Spearman's $\rho$. Green when sign matches expected direction with $|\rho|\geq0.4$ (or $|\rho|\leq0.2$ for stability tests).}
\label{tab:all_tests_rho}
\end{subtable}
\hfill
\begin{subtable}[t]{0.49\linewidth}
\centering
\footnotesize
\setlength{\tabcolsep}{2pt}
\begin{tabular}{lrrrrr}
\toprule
 & \multicolumn{3}{c}{\textit{Directional}} & \multicolumn{2}{c}{\textit{Stability}} \\
\cmidrule(lr){2-4} \cmidrule(lr){5-6}
Metric & DiffAud & Distract & Increm & Lengthen & Shorten \\
\midrule
\multicolumn{6}{l}{\textit{Lexical}} \\
\midrule
ROUGE-1 (F1) & \cellcolor{matchTrue}0.67 & \cellcolor{matchTrue}1.00 & \cellcolor{matchFalse}0.27 & \cellcolor{matchFalse}0.00 & \cellcolor{matchFalse}0.00 \\
ROUGE-2 (F1) & \cellcolor{matchTrue}0.67 & \cellcolor{matchTrue}0.80 & \cellcolor{matchFalse}0.20 & \cellcolor{matchFalse}0.00 & \cellcolor{matchFalse}0.00 \\
ROUGE-L (F1) & \cellcolor{matchTrue}0.67 & \cellcolor{matchTrue}1.00 & \cellcolor{matchFalse}0.20 & \cellcolor{matchFalse}0.00 & \cellcolor{matchFalse}0.00 \\
BLEU & \cellcolor{matchFalse}0.33 & \cellcolor{matchTrue}1.00 & \cellcolor{matchFalse}0.13 & \cellcolor{matchFalse}0.00 & \cellcolor{matchFalse}0.00 \\
chrF & \cellcolor{matchTrue}0.67 & \cellcolor{matchTrue}1.00 & \cellcolor{matchFalse}0.20 & \cellcolor{matchFalse}0.00 & \cellcolor{matchTrue}1.00 \\
Extr. Coverage & \cellcolor{matchTrue}0.67 & \cellcolor{matchTrue}0.80 & \cellcolor{matchTrue}0.53 & \cellcolor{matchFalse}0.00 & \cellcolor{matchFalse}0.00 \\
Extr. Density & \cellcolor{matchTrue}0.67 & \cellcolor{matchTrue}0.80 & \cellcolor{matchTrue}0.53 & \cellcolor{matchFalse}0.00 & \cellcolor{matchFalse}0.00 \\
Compression Ratio & \cellcolor{matchTrue}0.67 & \cellcolor{matchTrue}1.00 & \cellcolor{matchFalse}0.07 & \cellcolor{matchFalse}0.00 & \cellcolor{matchFalse}0.00 \\
Novel 1-grams (\%) & \cellcolor{matchFalse}0.33 & \cellcolor{matchFalse}0.00 & \cellcolor{matchTrue}0.71 & \cellcolor{matchFalse}0.00 & \cellcolor{matchFalse}0.00 \\
Novel 2-grams (\%) & \cellcolor{matchFalse}0.33 & \cellcolor{matchFalse}0.20 & \cellcolor{matchTrue}0.57 & \cellcolor{matchFalse}0.00 & \cellcolor{matchFalse}0.00 \\
Novel 3-grams (\%) & \cellcolor{matchFalse}0.33 & \cellcolor{matchFalse}0.20 & \cellcolor{matchTrue}0.57 & \cellcolor{matchFalse}0.00 & \cellcolor{matchFalse}0.00 \\
Summary Length & \cellcolor{matchTrue}0.67 & \cellcolor{matchFalse}0.00 & \cellcolor{matchTrue}1.00 & \cellcolor{matchFalse}0.00 & \cellcolor{matchFalse}0.00 \\
METEOR & \cellcolor{matchTrue}0.67 & \cellcolor{matchTrue}1.00 & \cellcolor{matchFalse}0.27 & \cellcolor{matchFalse}0.00 & \cellcolor{matchFalse}0.00 \\
Synt. Complexity (W) & \cellcolor{matchTrue}0.67 & \cellcolor{matchFalse}0.00 & \cellcolor{matchTrue}1.00 & \cellcolor{matchFalse}0.00 & \cellcolor{matchFalse}0.00 \\
Synt. Complexity (S) & \cellcolor{matchTrue}0.67 & \cellcolor{matchFalse}0.00 & \cellcolor{matchTrue}1.00 & \cellcolor{matchFalse}0.00 & \cellcolor{matchFalse}0.00 \\
\midrule
\multicolumn{6}{l}{\textit{Non-lexical}} \\
\midrule
BERTScore (F1) & \cellcolor{matchTrue}0.67 & \cellcolor{matchTrue}1.00 & \cellcolor{matchFalse}0.00 & \cellcolor{matchFalse}0.00 & \cellcolor{matchFalse}0.00 \\
BLANC & \cellcolor{matchFalse}0.33 & \cellcolor{matchFalse}0.00 & \cellcolor{matchFalse}0.07 & \cellcolor{matchTrue}1.00 & \cellcolor{matchTrue}1.00 \\
SummaQA (F1) & \cellcolor{matchFalse}0.00 & \cellcolor{matchFalse}0.00 & \cellcolor{matchFalse}0.00 & \cellcolor{matchTrue}1.00 & \cellcolor{matchTrue}1.00 \\
SUPERT & \cellcolor{matchTrue}0.50 & --- & --- & \cellcolor{matchTrue}1.00 & \cellcolor{matchTrue}1.00 \\
\midrule
\multicolumn{6}{l}{\textit{LLM judge: Llama-70B}} \\
\midrule
FActScore & \cellcolor{matchFalse}0.33 & \cellcolor{matchTrue}0.60 & \cellcolor{matchFalse}0.47 & \cellcolor{matchFalse}0.00 & \cellcolor{matchFalse}0.00 \\
Relevance & \cellcolor{matchFalse}0.00 & \cellcolor{matchFalse}0.40 & \cellcolor{matchFalse}0.07 & \cellcolor{matchTrue}1.00 & \cellcolor{matchTrue}1.00 \\
Coherence & \cellcolor{matchFalse}0.00 & \cellcolor{matchFalse}0.40 & \cellcolor{matchFalse}0.13 & \cellcolor{matchFalse}0.00 & \cellcolor{matchTrue}1.00 \\
Consistency & \cellcolor{matchFalse}0.00 & \cellcolor{matchFalse}0.20 & \cellcolor{matchFalse}0.07 & \cellcolor{matchTrue}1.00 & \cellcolor{matchTrue}1.00 \\
Fluency & \cellcolor{matchFalse}0.00 & \cellcolor{matchFalse}0.00 & \cellcolor{matchFalse}0.13 & \cellcolor{matchFalse}0.00 & \cellcolor{matchFalse}0.00 \\
Informativeness & \cellcolor{matchFalse}0.33 & \cellcolor{matchFalse}0.00 & \cellcolor{matchFalse}0.07 & \cellcolor{matchFalse}0.00 & \cellcolor{matchFalse}0.00 \\
Overall & \cellcolor{matchFalse}0.33 & \cellcolor{matchTrue}0.80 & \cellcolor{matchFalse}0.13 & \cellcolor{matchFalse}0.00 & \cellcolor{matchTrue}1.00 \\
Annotator & \cellcolor{matchFalse}0.33 & \cellcolor{matchFalse}0.00 & --- & \cellcolor{matchTrue}1.00 & \cellcolor{matchFalse}0.00 \\
Persona Precision & \cellcolor{matchFalse}0.00 & \cellcolor{matchFalse}0.00 & \cellcolor{matchFalse}0.00 & \cellcolor{matchTrue}1.00 & \cellcolor{matchTrue}1.00 \\
Persona Recall & \cellcolor{matchFalse}0.33 & \cellcolor{matchTrue}0.60 & \cellcolor{matchTrue}0.60 & \cellcolor{matchFalse}0.00 & \cellcolor{matchFalse}0.00 \\
\midrule
\multicolumn{6}{l}{\textit{LLM judge: Prometheus-7B}} \\
\midrule
FActScore & \cellcolor{matchFalse}0.00 & \cellcolor{matchTrue}0.75 & \cellcolor{matchFalse}0.38 & \cellcolor{matchFalse}0.00 & \cellcolor{matchFalse}0.00 \\
Relevance & \cellcolor{matchTrue}0.67 & \cellcolor{matchFalse}0.40 & \cellcolor{matchFalse}0.27 & \cellcolor{matchTrue}1.00 & \cellcolor{matchFalse}0.00 \\
Coherence & \cellcolor{matchFalse}0.33 & \cellcolor{matchFalse}0.40 & \cellcolor{matchTrue}0.53 & \cellcolor{matchFalse}0.00 & \cellcolor{matchFalse}0.00 \\
Consistency & \cellcolor{matchTrue}0.67 & \cellcolor{matchFalse}0.40 & \cellcolor{matchFalse}0.27 & \cellcolor{matchFalse}0.00 & \cellcolor{matchTrue}1.00 \\
Fluency & \cellcolor{matchFalse}0.33 & \cellcolor{matchTrue}0.60 & \cellcolor{matchFalse}0.47 & \cellcolor{matchFalse}0.00 & \cellcolor{matchFalse}0.00 \\
Informativeness & \cellcolor{matchFalse}0.33 & \cellcolor{matchFalse}0.20 & \cellcolor{matchFalse}0.40 & \cellcolor{matchFalse}0.00 & \cellcolor{matchFalse}0.00 \\
Overall & \cellcolor{matchFalse}0.33 & \cellcolor{matchTrue}0.60 & \cellcolor{matchFalse}0.47 & \cellcolor{matchFalse}0.00 & \cellcolor{matchFalse}0.00 \\
Annotator & \cellcolor{matchFalse}0.33 & \cellcolor{matchFalse}0.20 & \cellcolor{matchFalse}0.27 & \cellcolor{matchFalse}0.00 & \cellcolor{matchFalse}0.00 \\
Persona Precision & \cellcolor{matchTrue}0.67 & \cellcolor{matchTrue}0.80 & \cellcolor{matchFalse}0.43 & \cellcolor{matchTrue}1.00 & \cellcolor{matchFalse}0.00 \\
Persona Recall & \cellcolor{matchFalse}0.33 & \cellcolor{matchTrue}0.60 & \cellcolor{matchFalse}0.31 & \cellcolor{matchFalse}0.00 & \cellcolor{matchFalse}0.00 \\
\bottomrule
\end{tabular}
\caption{Match-rate $M$ (fraction of instances where metric responds as expected). Green when $M \geq 0.5$.}
\label{tab:all_tests_m}
\end{subtable}
\caption{Spearman's $\rho$ (left) and match-rate $M$ (right) across all perturbation tests.}
\label{tab:all_tests_combined}
\end{table*}

%% file: 6-Human-Evaluation-Results.tex
\input{tables/metric-agreement}

\section{Human Eval Results \hyperref[chsevenrq2]{\chsevenrqtwo}}\label{sec:ch-7-human-eval-results}

We collected 140 completed query sessions from 20 annotators, yielding 420 pairwise comparisons across the three-match tournament described in~\Cref{sec:ch-7-human-eval-setup}. On average, annotators spent roughly four minutes per page (median 258s). Sixteen annotators completed all eight assigned queries. For reference-based metrics, we concatenate the titles of the retrieved papers as the reference. For metrics that require the source, we concatenate the abstracts of the retrieved papers.

\textit{DeepSeek-V3.1 is preferred over Llama-3.3-70B.}
Aggregating across both prompting conditions and all tournament rounds, DeepSeek-V3.1 wins 72.8\% of its matchups (91/125) in the final round, against 27.2\% for Llama-3.3-70B-Instruct. The gap is even more pronounced in direct cross-model first-round matchups, where DeepSeek wins 73.1\% of head-to-head comparisons (68/93). This preference is consistent across the four-variant breakdown shown in \Cref{tab:variant-winrates}. Both DeepSeek variants outperform both Llama variants in both rounds.

\textit{Personalization is preferred when generated by a stronger model.}
The effect of conditioning on the persona profile depends on where in the tournament the comparison occurs (\Cref{tab:personalization}). In the final round, annotators prefer the personalized summary 63.2\% of the time (79/125) over the generic summary, with 15 ties (``neither''). The first-round picture is less clean: in the \emph{A vs.\ B} bracket, personalized summaries win only 46.5\% of the time (60/129) while in the \emph{C vs.\ D} bracket, personalized summaries win 62.2\% (79/127), matching the final-round rate. We hypothesize this effect is a result of the fact that the first-round result aggregates across both models, and our cross-model comparisons are dominated by a strong preference for DeepSeek-V3.1 over Llama-3.3-70B (\Cref{tab:variant-winrates}). If Llama is simply less effective at personalization, then including its personalized outputs in the first-round average would dilute the personalization signal there, while the final round, which is disproportionately populated by the stronger model's outputs, would surface it more cleanly.

\input{tables/variant-win-rates}

To test this hypothesis while controlling for model strength, we restricted the data to comparisons in which one summary was personalized and the other generic, excluding same-condition matchups and ``neither'' picks. Each remaining comparison (N = 248, from 19 annotators) contributed one observation with a binary outcome indicating whether the personalized side won. We then fit a logistic regression of \texttt{personalized\_won} on three categorical predictors: the model on the personalized side (\texttt{model\_pq}), the model on the generic side (\texttt{model\_generic}), and the round in which the comparison occurred (\texttt{round\_stage}: round 1 or final), with DeepSeek-V3.1 and the final round as reference levels. Because each annotator contributed multiple comparisons, we computed cluster-robust standard errors grouped by annotator~\cite{MACKINNON2023272}. Under this parameterization, the intercept estimates the log-odds of the personalized variant winning at the reference levels (personalized DeepSeek vs.\ generic DeepSeek in the final round), and thus, after exponentiation to an odds ratio, directly quantifies the personalization effect.

We report the results of the regression analysis in~\Cref{tab:personalization-regression}, which support the model-strength hypothesis. The personalized DeepSeek summary is preferred over the generic DeepSeek summary with an odds ratio of $1.86$ (95\% CI $[1.01, 3.42]$, $p = 0.046$). The penalty for putting Llama on the personalized side is large and significant: $\text{OR} = 0.43$ ($[0.22, 0.86]$, $p = 0.016$), meaning that swapping the personalized model from DeepSeek to Llama more than halves the odds that the personalized side wins. The coefficient on the generic-side model trends in the expected complementary direction ($\text{OR} = 1.69$, $[0.95, 3.01]$, $p = 0.076$): personalization wins more often when the generic opponent is Llama, though this effect is not significant at $\alpha = 0.05$. The round-stage coefficient is small and non-significant ($\text{OR} = 0.82$, $[0.44, 1.53]$, $p = 0.53$), indicating that once model identity is controlled for, the apparent round-level differences in raw win rates largely vanish. This is consistent with the claim that those differences were driven by the composition of models in each round rather than by the round itself.

\textit{Automatic and LLM-based metrics correlate poorly with human judgment.}
To assess whether existing summarization metrics recover the human preferences just described, we compute Krippendorff's $\alpha$ between each metric and the annotator's choice on every comparison in our evaluation~\cite{Krippendorff2004MeasuringTR}. A metric agrees with the annotator on a given pair if it assigns a higher score to the summary that the annotator selected, corresponding to $\alpha\approx0$. We count the metric's decision as ``neither'' if the scores are within a per-metric threshold of similarity, reported in the appendix.

\Cref{tab:metric-agreement-alpha} reports Krippendorff's $\alpha$ grouped into three families: reference-based and reference-free automatic metrics (e.g., ROUGE, BERTScore, BLANC), LLM-as-a-judge scores using Prometheus-7B as the evaluator, and LLM-as-a-judge scores using Llama-3.3-70B-Instruct as the evaluator. Across all three families, chance-corrected agreement remains at or near chance ($\alpha = 0$): automatic metrics range from $-0.029$ to $0.173$ (mean $0.049$), Prometheus-based judgments range from $-0.150$ to $0.158$ (mean $-0.022$), and Llama-based judgments range from $-0.179$ to $0.053$ (mean $-0.087$), with all but FactScore ($0.053$) and Fluency ($0.000$) falling below chance. Notably, the persona-aware variants we designed specifically to capture informational satisfaction, \textsc{persona\_precision} and \textsc{persona\_recall}, are among the worst performers under the Prometheus judge ($\alpha = -0.150$ for both) and remain at or below chance under Llama, with $\alpha$ values across the two judges ranging from $-0.150$ to $-0.087$, indicating systematic disagreement with annotator preferences rather than mere noise.

The strongest performers are reference-based metrics anchored to a reference summary (SummaQA F1 at $\alpha = 0.173$, ROUGE-2 F1 at $0.117$, ROUGE-1 F1 at $0.077$), but even these correspond to only slight agreement on conventional interpretations of $\alpha$ and fall far short of the level required to use them as proxies for personalized informational satisfaction. The LLM-judge protocols, including those explicitly conditioned on the annotator's persona profile, perform no better than reference-free automatic metrics. Together, these results support the central claim of this paper: existing summarization metrics, including state-of-the-art LLM-based judges, are insufficient measures of how well a summary serves an individual reader's informational needs.

%% file: tables/metric-agreement.tex
\begin{table*}[t!]
  \centering
  \begin{subtable}[t]{0.3\linewidth}
    \centering
    \begin{tabular}{lc}
      \toprule
      \textbf{Metric} & \textbf{$\alpha$} \\
      \midrule
      ROUGE-1 F1        & 0.077 \\
      ROUGE-2 F1        & 0.117 \\
      ROUGE-L F1        & 0.032 \\
      chrF              & 0.069 \\
      SummaQA F1        & \textbf{0.173} \\
      BERTScore F1      & -0.029 \\
      BLEU              & 0.024 \\
      METEOR            & 0.018 \\
      BLANC             & 0.018 \\
      SUPERT            & -0.005 \\
      \bottomrule
    \end{tabular}
    \caption{Traditional metrics.}
    \label{tab:metric-agreement-alpha-traditional}
  \end{subtable}
  \hfill
  \begin{subtable}[t]{0.32\linewidth}
    \centering
    \begin{tabular}{lc}
          &  \\
      \toprule
      \textbf{Metric} & \textbf{$\alpha$} \\
      \midrule
      Overall                  & -0.036 \\
      Relevance                & 0.001 \\
      Coherence                & \textbf{0.158} \\
      Consistency              & 0.027 \\
      Fluency                  & -0.122 \\
      Informativeness          & 0.083 \\
      FactScore                & -0.009 \\
      Persona-precision        & -0.150 \\
      Persona-recall           & -0.150 \\
      \bottomrule
    \end{tabular}
    \caption{LLM-as-a-judge (Prometheus-7B).}
    \label{tab:metric-agreement-alpha-prometheus}
  \end{subtable}
  \hfill
  \begin{subtable}[t]{0.35\linewidth}
    \centering
    \begin{tabular}{lc}
      &  \\
      \toprule
      \textbf{Metric} & \textbf{$\alpha$} \\
      \midrule
      Overall            & -0.103 \\
      Relevance          & -0.153 \\
      Coherence          & -0.106 \\
      Consistency        & -0.102 \\
      Fluency            & 0.000 \\
      Informativeness    & -0.179 \\
      FactScore          & \textbf{0.053} \\
      Persona-precision  & -0.105 \\
      Persona-recall     & -0.087 \\      \bottomrule
    \end{tabular}
    \caption{LLM-as-a-judge (Llama-3.3-70B-Instruct).}
    \label{tab:metric-agreement-alpha-llama}
  \end{subtable}

  \caption{$\alpha$ between metrics and human pairwise preferences. Chance agreement corresponds to $\alpha = 0$. No metric family substantively exceeds chance, including persona-aware variants designed to capture informational satisfaction.}
  \label{tab:metric-agreement-alpha}
\end{table*}

%% file: tables/variant-win-rates.tex
\begin{table*}[t]
\centering

\begin{subtable}[t]{0.54\textwidth}
\centering
\small
\begin{tabular}{lcccccc}
& \multicolumn{3}{c}{\textbf{Round 1}} & \multicolumn{3}{c}{\textbf{Final}} \\
\cmidrule(lr){2-4}\cmidrule(lr){5-7}
\textbf{Variant} & W/N & \% & $p$ & W/N & \% & $p$ \\
\midrule
DeepSeek-V3.1 (pers.) & 88/140 & 62.9 & $<$0.001 & 58/88 & \textbf{65.9} & 0.002 \\
DeepSeek-V3.1 (gen.)      & 75/140 & 53.6 & 0.010 & 33/75 & 44.0 & 0.474 \\
Llama-3.3-70B (pers.) & 51/140 & 36.4 & 0.066 & 21/51 & 41.2 & 0.284 \\
Llama-3.3-70B (gen.)      & 42/140 & 30.0 & $<$0.001 & 13/42 & 31.0 & 0.112 \\
\end{tabular}
\vspace{2mm}
\caption{Win rates for each of the four summary variants in round 1 (out of 140 round-1 matchups per variant) and in the final round (out of the variant's round-1 wins). We compare each model for both the generic (gen.) and personalized (pers.) settings. DeepSeek-V3.1 with persona conditioning is the most preferred variant across both rounds. The $p$ columns report annotator-clustered bootstrap significance against a 50/50 null (two-sided, decided matchups only).}
\label{tab:variant-winrates}
\end{subtable}
\hfill
\begin{subtable}[t]{0.44\textwidth}
\centering
\small
\begin{tabular}{lcccc}
 & & & & \\
\textbf{Comparison} & \textbf{Pers.} & \textbf{Gen.} & \textbf{Tie} & \textbf{$p$} \\
\midrule
Round 1 (A vs.\ B) & 60 (46.5\%) & 69 (53.5\%) & 11 & 0.408 \\
Round 1 (C vs.\ D) & 79 (62.2\%) & 48 (37.8\%) & 13 & \textbf{0.015} \\
Final round        & \textbf{79 (63.2\%)} & 46 (36.8\%) & 15 & 0.026 \\
\hdash{1}{5}
Same-model R1 & 51 (57.3\%) & 38 (42.7\%) & --- & 0.239 \\
\end{tabular}
\vspace{2mm}
\caption{Annotator preferences between persona-conditioned and generic summaries, by tournament position. Personalization is preferred in the final round and in the second first-round bracket (C vs.\ D), and---when model identity is held fixed---in same-model first-round matchups. The $p$ column reports annotator-clustered bootstrap significance against a 50/50 null; bold indicates $p < 0.05$.}
\label{tab:personalization}
\end{subtable}

\vspace{1.5ex}

\begin{subtable}[t]{\textwidth}
\centering
\small
\begin{tabular}{lrrrrr}
\textbf{Term} & \textbf{Coef.} & \textbf{SE} & \textbf{$z$} & \textbf{$p$} & \textbf{OR [95\% CI]} \\
\midrule
Intercept                              &  0.619 & 0.311 &  1.99 & 0.046 & 1.86 [1.01, 3.42] \\
\texttt{model\_pq} = Llama-3.3-70B     & -0.833 & 0.347 & -2.40 & 0.016 & 0.43 [0.22, 0.86] \\
\texttt{model\_generic} = Llama-3.3-70B &  0.523 & 0.295 &  1.77 & 0.076 & 1.69 [0.95, 3.01] \\
\texttt{round\_stage} = round 1        & -0.199 & 0.319 & -0.62 & 0.533 & 0.82 [0.44, 1.53] \\
\end{tabular}
\vspace{2mm}
\caption{Logistic regression of personalized-side win on personalized-side model, generic-side model, and tournament round. Reference levels: \texttt{model\_pq} = DeepSeek-V3.1, \texttt{model\_generic} = DeepSeek-V3.1, \texttt{round\_stage} = final. Standard errors are cluster-robust by annotator. $N = 248$ comparisons; personalized side won 58.5\% of comparisons overall.}
\label{tab:personalization-regression}
\end{subtable}

\caption{Human evaluation results. (\subref{tab:variant-winrates}) Per-variant win rates across the $2 \times 2$ design. (\subref{tab:personalization}) Aggregated personalized-vs-generic preferences by tournament position. (\subref{tab:personalization-regression}) Logistic regression of the personalized-side win with cluster-robust standard errors.}
\label{tab:ch-7-human-eval}
\end{table*}

%% file: 7-Conclusion.tex
\section{Conclusion}

 We introduced \textit{information satisfaction} as a user-centered axis of summarization evaluation and asked whether existing metrics can measure it. The answer is no: Our perturbation tests reveal that traditional metrics, embedding-based metrics, and even strong LLM-as-judge protocols fail basic robustness checks. Most notably, nearly every metric we evaluated is insensitive to (or even \emph{anticorrelated} with) audience-shift rewrites—the very dimension along which informational satisfaction varies between users. Our expert human evaluation confirms the consequence, as no metric family exceeds chance agreement with reader preferences anchored to a specific query and persona. 
Progress on user-centered summarization will therefore require evaluation frameworks that explicitly model who the summary is for and what they seek to learn. Conditioning on the persona is not by itself sufficient, however. Our persona metric variants incorporate the reader's role, domain, and information needs, yet do not consistently pass the perturbation or human correlation tests. Their failure indicates that the bottleneck is not the absence of persona information but how satisfaction is measured. Both metrics reduce satisfaction to the presence or absence of nuggets that an LLM judges relevant. This assumes the model's estimate of a reader's needs matches the reader's own, and it treats every information requirement as equally and independently satisfiable. A metric that truly measures information satisfaction will instead need to weight information by its importance to the specific reader, account for what that reader already knows, and capture the comparative judgment a person makes when one adequate summary serves them better than another. Our analysis and human-annotated dataset provide a foundation for future research on evaluation of information satisfaction.

%% file: 8-appendix.tex



\clearpage
\input{tables/thresholds}
\clearpage
\input{tables/prompts}
\clearpage

%% file: tables/thresholds.tex
%
\begin{table*}[h]
  \centering
  \small

  \begin{subtable}[t]{0.48\linewidth}
    \centering
    \begin{tabular}{lcl}
    & & \\
    & & \\
    & & \\
      \toprule
      \textbf{Metric} & \textbf{Thr.} & \textbf{Range} \\
      \midrule
      \multicolumn{3}{l}{\textit{Overlap-based}} \\
      \midrule
      ROUGE-1 (P / R / F1)            & 0.05 & $[0, 1]$ \\
      ROUGE-2 (P / R / F1)            & 0.03 & $[0, 1]$ \\
      ROUGE-L (P / R / F1)            & 0.05 & $[0, 1]$ \\
      BLEU                            & 2.0  & $[0, 100]$ \\
      chrF++                          & 2.0  & $[0, 100]$ \\
      METEOR                          & 0.05 & $[0, 1]$ \\
      CIDEr                           & 0.05 & $[0, {\sim}10]$ \\
      \midrule
      \multicolumn{3}{l}{\textit{DataStats (surface)}} \\
      \midrule
      Extractive coverage             & 0.05 & $[0, 1]$ \\
      Extractive density              & 0.1  & $[0, {\sim}\infty)$ \\
      Compression ratio               & 0.1  & $[0, {\sim}\infty)$ \\
      Summary length (words)          & 5.0  & $[0, \infty)$ \\
      \% novel 1-grams                & 0.05 & $[0, 1]$ \\
      \% novel 2-grams                & 0.05 & $[0, 1]$ \\
      \% novel 3-grams                & 0.05 & $[0, 1]$ \\
      \% rep.\ 1-grams in summary     & 0.01 & $[0, 1]$ \\
      \% rep.\ 2-grams in summary     & 0.01 & $[0, 1]$ \\
      \% rep.\ 3-grams in summary     & 0.01 & $[0, 1]$ \\
      \midrule
      \multicolumn{3}{l}{\textit{Pairwise length}} \\
      \midrule
      Length (word-count diff.)       & 5.0  & $[0, \infty)$ \\
      \bottomrule
    \end{tabular}
    \caption{Lexical metrics.}
    \label{tab:thresholds-lexical}

    \vspace{1em}

    \begin{tabular}{lcl}
    & & \\
    & & \\
    & & \\
    & & \\
    & & \\
      \toprule
      \textbf{Metric} & \textbf{Thr.} & \textbf{Range} \\
      \midrule
      BERTScore (P / R / F1)          & 0.02 & $[0, 1]$ \\
      SUPERT                          & 0.05 & $[0, 1]$ \\
      SummaQA (avg.\ prob.)           & 0.05 & $[0, 1]$ \\
      SummaQA (avg.\ F-score)         & 0.05 & $[0, 1]$ \\
      BLANC                           & 0.02 & $[-1, 1]$ \\
      \bottomrule
    \end{tabular}
    \caption{Non-lexical (embedding- and model-based) metrics.}
    \label{tab:thresholds-nonlexical}
  \end{subtable}
  \hfill
  \begin{subtable}[t]{0.48\linewidth}
    \centering
    \begin{tabular}{lcl}
      \toprule
      \textbf{Metric} & \textbf{Thr.} & \textbf{Range} \\
      \midrule
      \multicolumn{3}{l}{\textit{Raw counts}} \\
      \midrule
      Words                           & 10.0 & $[0, \infty)$ \\
      Sentences                       & 1.0  & $[0, \infty)$ \\
      Verb phrases                    & 2.0  & $[0, \infty)$ \\
      Clauses                         & 2.0  & $[0, \infty)$ \\
      T-units                         & 1.0  & $[0, \infty)$ \\
      Dependent clauses               & 1.0  & $[0, \infty)$ \\
      Complex T-units                 & 1.0  & $[0, \infty)$ \\
      Coordinate phrases              & 1.0  & $[0, \infty)$ \\
      Complex nominals                & 2.0  & $[0, \infty)$ \\
      \midrule
      \multicolumn{3}{l}{\textit{Ratios}} \\
      \midrule
      Words / sentence                & 3.0  & $[0, \infty)$ \\
      Words / T-unit                  & 3.0  & $[0, \infty)$ \\
      Words / clause                  & 2.0  & $[0, \infty)$ \\
      Clauses / sentence              & 0.5  & $[0, \infty)$ \\
      Verb phrases / T-unit           & 0.5  & $[0, \infty)$ \\
      Clauses / T-unit                & 0.5  & $[0, \infty)$ \\
      Dep.\ clauses / clause          & 0.1  & $[0, \infty)$ \\
      Dep.\ clauses / T-unit          & 0.2  & $[0, \infty)$ \\
      T-units / sentence              & 0.3  & $[0, \infty)$ \\
      Complex T-units / T-unit        & 0.2  & $[0, \infty)$ \\
      Coord.\ phrases / T-unit        & 0.2  & $[0, \infty)$ \\
      Coord.\ phrases / clause        & 0.1  & $[0, \infty)$ \\
      Complex nominals / T-unit       & 0.5  & $[0, \infty)$ \\
      Complex nominals / clause       & 0.3  & $[0, \infty)$ \\
      \bottomrule
    \end{tabular}
    \caption{Syntactic complexity metrics.}
    \label{tab:thresholds-syntactic}

    \vspace{1em}

    \begin{tabular}{lcl}
      \toprule
      \textbf{Metric} & \textbf{Thr.} & \textbf{Range} \\
      \midrule
      \multicolumn{3}{l}{\textit{LLM Judge (per-dimension)}} \\
      \midrule
      Relevance                       & 0.5  & $[1, 5]$ \\
      Coherence                       & 0.5  & $[1, 5]$ \\
      Consistency                     & 0.5  & $[1, 5]$ \\
      Fluency                         & 0.5  & $[1, 5]$ \\
      Informativeness                 & 0.5  & $[1, 5]$ \\
      Overall                         & 0.5  & $[1, 5]$ \\
      \midrule
      \multicolumn{3}{l}{\textit{Other}} \\
      \midrule
      FActScore                       & 0.1  & $[0, 1]$ \\
      \bottomrule
    \end{tabular}
    \caption{LLM-based metrics.}
    \label{tab:thresholds-llm}
  \end{subtable}
    \caption{Per-metric similarity thresholds used in the agreement
  analysis (\Cref{sec:ch-7-human-eval-results,tab:metric-agreement-alpha}). When $|\text{metric}(A) - \text{metric}(B)|$
  is below the listed threshold, the metric's decision is counted as
  ``Neither.'' Thresholds reflect the typical scale and variance of each
  metric.}
  \label{tab:thresholds}
\end{table*}

%% file: tables/prompts.tex
\begin{table*}[h]
\centering
\small
\begin{tabular}{p{.95\linewidth}}
\toprule
\textbf{Prompt: LLM Judge} \\
\midrule
\end{tabular}
\begin{lstlisting}[style=promptstyle]
###Task Description:
An instruction (might include an Input inside it), a response to evaluate, and a score rubric representing an evaluation criteria are given.
1. Write a detailed feedback that assesses the quality of the response strictly based on the given score rubric, not evaluating in general.
2. After writing a feedback, write a score that is an integer between 1 and 5. You should refer to the score rubric.
3. The output format should look as follows: "Feedback: (write a feedback for criteria) [RESULT] (an integer number between 1 and 5)"
4. Please do not generate any other opening, closing, or explanations.

###Instruction:
You are evaluating a summary from the perspective of a specific person. Consider their background and information needs when assessing quality.

Annotator profile:
- Role: {role}
- Domain: {domain}
- Information needs: {info_needs}
- Query: {query}

Evaluate the quality of the following summary on the dimension of {dimension}, from this person's perspective.

Summary:
{summary}

###Response to evaluate:
{summary}

###Score Rubric:
{rubric}
\end{lstlisting}
\begin{tabular}{p{\textwidth}}
\bottomrule
\end{tabular}
\caption{Prompt template for LLM Judges. Score rubrics in~\Cref{tab:ch-7-llm-judge-rubrics}. Source is also provided in prompt for consistency and relevance.}
\label{tab:ch-7-prompt-llm-judge}
\end{table*}

\begin{table*}[h]
\centering
\small
\begin{tabular}{p{.95\textwidth}}
\toprule
\textbf{Score Rubrics} \\
\midrule
\end{tabular}

    \begin{tabular}{p{.9\textwidth}}
        \texttt{Coherence} \\ \hdash{1}{1} 
    \end{tabular}
    \begin{lstlisting}[style=promptstyle]
Is the summary well-organized and easy to read?
[1] The summary is incoherent and very difficult to follow.
[2] The summary has poor organization with frequent disjointed transitions.
[3] The summary is somewhat organized but has noticeable structural issues.
[4] The summary is well-organized with only minor flow issues.
[5] The summary is excellently structured and flows naturally throughout.
    \end{lstlisting}

    \begin{tabular}{p{.9\textwidth}}
        \texttt{Consistency} \\ \hdash{1}{1} 
    \end{tabular}
    \begin{lstlisting}[style=promptstyle]
Is the summary factually consistent with the source document?
[1] The summary contains major factual errors or hallucinations not found in the source.
[2] The summary contains several factual inconsistencies with the source.
[3] The summary is mostly consistent but contains a few minor factual errors.
[4] The summary is largely consistent with only negligible inaccuracies.
[5] All information in the summary is fully consistent with the source document.
    \end{lstlisting}

    \begin{tabular}{p{.9\textwidth}}
        \texttt{Fluency} \\ \hdash{1}{1} 
    \end{tabular}
    \begin{lstlisting}[style=promptstyle]
Is the summary grammatically correct and well-written?
[1] The summary has many grammatical errors and is poorly written.
[2] The summary has frequent grammatical issues that hinder readability.
[3] The summary has occasional grammatical errors but is generally readable.
[4] The summary is well-written with only minor grammatical issues.
[5] The summary has flawless grammar and excellent writing quality.
    \end{lstlisting}

    \begin{tabular}{p{.9\textwidth}}
        \texttt{Relevance} \\ \hdash{1}{1} 
    \end{tabular}
    \begin{lstlisting}[style=promptstyle]
Does the summary capture the key information from the source?
[1] The summary is completely irrelevant and fails to capture any key information from the source document.
[2] The summary captures very little key information and misses most important points.
[3] The summary captures some key information but misses several important points.
[4] The summary captures most key information with only minor omissions.
[5] The summary captures all key points and essential information from the source.
    \end{lstlisting}

    \begin{tabular}{p{.9\textwidth}}
        \texttt{Informativeness} \\ \hdash{1}{1}
    \end{tabular}
    \begin{lstlisting}[style=promptstyle]
How useful and informative is the summary to this specific person, given their background and information needs?
    
[1] The summary provides no useful information for this person's specific needs and background. It fails to address their domain or information requirements.
[2] The summary provides very little information relevant to this person's needs. It largely misses content that would be useful given their role and domain expertise.
[3] The summary provides some information relevant to this person's needs but lacks important domain-specific details or fails to address their stated information requirements.
[4] The summary is informative for this person's needs, covering most relevant details for their role and domain, with only minor gaps in addressing their information requirements.
[5] The summary is highly informative for this person's specific needs, thoroughly addressing their domain expertise and information requirements with relevant detail and appropriate context. 
    \end{lstlisting}
    
\begin{tabular}{p{\textwidth}}
\bottomrule
\end{tabular}
\caption{Score rubrics for prompt in~\Cref{tab:ch-7-prompt-llm-judge}.}
\label{tab:ch-7-llm-judge-rubrics}
\end{table*}


\begin{table}[h]
\centering
\small
\begin{tabular}{p{.9\columnwidth}}
\toprule
\textbf{Prompt: LLM Judge Annotator} \\
\midrule
\end{tabular}
\begin{lstlisting}[style=promptstyle]
###Task Description:
You are evaluating a summary from the perspective of a specific annotator. Consider their background and information needs when assessing quality.
1. Write a detailed feedback that assesses strictly whether the summary best addresses the annotator's query, not the fidelity or the style.
2. After writing a feedback, write a score that is an integer between 1 and 5. You should refer to the score rubric.
3. The output format should look as follows: "Feedback: (write a feedback for criteria) [RESULT] (an integer number between 1 and 5)"
4. Please do not generate any other opening, closing, or explanations.

###Introduction
Automatic summarization tools condense large amounts of text into shorter versions, highlighting the most important points.
But "important" is subjective - what matters to one reader might not matter to another.
This study explores what individual readers actually want from summaries of scientific papers, and how we can measure whether a summary was useful to them.
**We're focused on whether the summary answers your query, not the style or fidelity of the summary.**

###Annotator profile:
- Role: {role}
- Domain: {domain}
- Information needs: {info_needs}

###Query:
{query}

###Response to evaluate:
{summary}

###Score Rubric:
How well does the summary address the annotator's specific query, given their background and information needs?
        
[1] The summary does not address the annotator's query at all. It provides no information that would help answer what they asked about.
[2] The summary barely addresses the query, touching on it only tangentially or providing very little of the requested information.
[3] The summary partially addresses the query, providing some relevant information but missing important aspects of what was asked.
[4] The summary addresses the query well, covering most of what was asked with only minor gaps.
[5] The summary fully and directly addresses the annotator's query, providing the requested information clearly and completely.
\end{lstlisting}
\begin{tabular}{p{\textwidth}}
\bottomrule
\end{tabular}
\caption{Prompt template for LLM Judge Annotator, in which the prompt reflects the instructions given to the human annotators. Score rubrics in~\Cref{tab:ch-7-llm-judge-rubrics}. Source is also provided in prompt for consistency and relevance.}
\label{tab:ch-7-prompt-llm-judge-annotator}
\end{table}


\begin{table}[h]
\centering
\small
\begin{tabular}{p{.95\columnwidth}}
\toprule
\textbf{Prompt: Persona Precision} \\
\midrule
\end{tabular}

    \begin{tabular}{p{.9\columnwidth}}
        \texttt{Step: Extract} \\ \hdash{1}{1} 
    \end{tabular}
    \begin{lstlisting}[style=promptstyle]
Break down the following summary into a list of independent informational nuggets. Each nugget should be a single, self-contained piece of information -- a minimal, atomic statement that conveys one fact or claim.

## Summary
{summary}

## Instructions
List each informational nugget on its own line, prefixed with "- ". Output ONLY the list of nuggets, nothing else.

Example format:
- The study used a randomized controlled trial design.
- The sample included 500 participants from three hospitals.
- Treatment group showed 23% improvement over control.
    \end{lstlisting}

    \begin{tabular}{p{.9\columnwidth}}
    \texttt{Step: Relevance} \\ \hdash{1}{1} 
    \end{tabular}
    \begin{lstlisting}[style=promptstyle]
Determine whether the following informational nugget from a summary is relevant to the given persona and their query.

A nugget is "relevant" if it addresses the persona's information needs, relates to their domain of expertise, or helps answer their query. A nugget is "not relevant" if it is unrelated to what this persona would care about.

## Persona
- Role: {role}
- Domain: {domain}
- Information needs: {info_needs}

## Query
{query}

## Nugget
{nugget}

## Instructions
Is this nugget relevant to the persona's needs and query? Respond with ONLY "relevant" or "not relevant".
    \end{lstlisting}

\begin{tabular}{p{\columnwidth}}
\bottomrule
\end{tabular}
\caption{Prompts for Persona Precision metric.}
\label{tab:ch-7-persona-precision-prompt}
\end{table}


\begin{table}[h]
\centering
\small
\begin{tabular}{p{.95\columnwidth}}
\toprule
\textbf{Prompt: Persona Recall} \\
\midrule
\end{tabular}

    \begin{tabular}{p{.9\columnwidth}}
        \texttt{Step: Extract} \\ \hdash{1}{1} 
    \end{tabular}
    \begin{lstlisting}[style=promptstyle]
Break down the following summary into a list of independent informational nuggets. Each nugget should be a single, self-contained piece of information -- a minimal, atomic statement that conveys one fact or claim.

## Summary
{summary}

## Instructions
List each informational nugget on its own line, prefixed with "- ". Output ONLY the list of nuggets, nothing else.

Example format:
- The study used a randomized controlled trial design.
- The sample included 500 participants from three hospitals.
- Treatment group showed 23% improvement over control.
    \end{lstlisting}

    \begin{tabular}{p{.9\columnwidth}}
    \texttt{Step: Requirements} \\ \hdash{1}{1}
    \end{tabular}
    \begin{lstlisting}[style=promptstyle]
You are given a persona description and their query about a source document. Generate a list of informational nugget requirements -- specific pieces of information that a good summary SHOULD contain to serve this persona's needs.

Each requirement should be an atomic, specific piece of information that this persona would expect or need from a summary of the source document.

## Persona
- Role: {role}
- Domain: {domain}
- Information needs: {info_needs}

## Query
{query}

## Source Document
{source}

## Instructions
Based on the persona's background, information needs, and query, list the key informational nuggets that a good summary should include. Each nugget should be a single, specific piece of information from the source document.

List each requirement on its own line, prefixed with "- ". Output ONLY the list of requirements, nothing else.

Example format:
- The main methodology used in the study.
- Key quantitative results and effect sizes.
- Limitations that affect practical application.
    \end{lstlisting}

    \begin{tabular}{p{.9\columnwidth}}
    \texttt{Step: Coverage} \\ \hdash{1}{1}
    \end{tabular}
    \begin{lstlisting}[style=promptstyle]
Determine whether the following informational requirement is covered by any of the summary nuggets below.

A requirement is "covered" if any of the summary nuggets address the same information, even if phrased differently.

## Requirement
{requirement}

## Summary Nuggets
{nuggets}

## Instructions
Is the requirement addressed by any of the summary nuggets above? Respond with ONLY "covered" or "not covered".
    \end{lstlisting}

\begin{tabular}{p{\columnwidth}}
\bottomrule
\end{tabular}
\caption{Prompts for Persona Recall metric.}
\label{tab:ch-7-persona-recall-prompt}
\end{table}


\begin{table}[h]
\centering
\small
\begin{tabular}{p{.95\columnwidth}}
\toprule
\textbf{Prompt: Perturbation Prompts} \\
\midrule
\end{tabular}

    \begin{tabular}{p{.9\columnwidth}}
        \texttt{Test: Lengthen} \\ \hdash{1}{1}
    \end{tabular}
    \begin{lstlisting}[style=promptstyle]
You are given a summary and the source document it was based on.

Rewrite the summary to be longer and more detailed. You MUST follow these rules:
- Do NOT add any new information that is not already present in or directly inferable from the original summary.
- Keep all original facts, claims, and findings intact.
- Expand by elaborating on existing points, adding transitional phrases, and using more descriptive language.
- The rewritten summary should be approximately 50% longer than the original.

Source document:
{source}

Original summary:
{summary}

Rewritten (longer) summary:
    \end{lstlisting}

    \begin{tabular}{p{.9\columnwidth}}
    \texttt{Test: Shorten} \\ \hdash{1}{1}
    \end{tabular}
    \begin{lstlisting}[style=promptstyle]
You are given a summary and the source document it was based on.

Rewrite the summary to be shorter and more concise. You MUST follow these rules:
- Do NOT remove any key information, facts, claims, or findings from the original summary.
- Preserve all substantive content while using fewer words.
- Remove redundancy, simplify phrasing, and tighten sentence structure.
- The rewritten summary should be approximately 50% shorter than the original.

Source document:
{source}

Original summary:
{summary}

Rewritten (shorter) summary:
    \end{lstlisting}

    \begin{tabular}{p{.9\columnwidth}}
    \texttt{Test: Different Audience} \\ \hdash{1}{1}
    \end{tabular}
    \begin{lstlisting}[style=promptstyle]
You are given a summary of a source document. The summary was originally written for: {original_audience}.

Rewrite this summary for a different target audience: {target_audience}.

You should:
- Adjust the language, terminology, and level of detail to be appropriate for the target audience.
- You may change which information is emphasized or how concepts are explained.
- Base your rewrite on the source document, not just the original summary.
- The rewrite should be roughly the same length as the original summary.

Source document:
{source}

Original summary (written for {original_audience}):
{summary}

Rewritten summary (for {target_audience}):
    \end{lstlisting}

\begin{tabular}{p{\columnwidth}}
\bottomrule
\end{tabular}
\caption{Prompts for the synthetic perturbations tests, described in~\Cref{sec:ch-7-experimental-setup}.}
\label{tab:ch-7-perturbation-prompt}
\end{table}